\documentclass[letterpaper, 10 pt, conference]{ieeeconf}  

\IEEEoverridecommandlockouts                              

\usepackage{amsmath}
\usepackage{amssymb}
\usepackage{array}
\usepackage{mathtools}

\usepackage{ntheorem}
\theorembodyfont{\normalfont}  
\theoremheaderfont{\bfseries}  

\usepackage{graphicx}
\usepackage{epstopdf} 
\usepackage[caption=false,font=footnotesize]{subfig}

\usepackage{textcomp}
\usepackage{verbatim}
\usepackage{float}
\usepackage{booktabs}
\usepackage{multirow}

\usepackage{cite}
\usepackage{url} 

\usepackage{algorithm}
\usepackage{algorithmic} 

\makeatletter
\let\NAT@parse\undefined 
\makeatother
\usepackage[colorlinks=true,
            linkcolor=green,
            citecolor=blue,
            urlcolor=blue]{hyperref} 

\title{\LARGE \bf
DA-VPC: Disturbance-Aware Visual Predictive Control Scheme of Docking Maneuvers for Autonomous Trolley Collection
}
\author{Yuhan~Pang$^{1\dag}$, 
        Bingyi~Xia$^{1\dag}$, 
        Zhe~Zhang$^1$,
        Zhirui~Sun$^1$,
        Peijia~Xie$^1$,
        Bike~Zhu$^1$, 
        Wenjun~Xu$^2$, 
        Jiankun~Wang$^1$
\thanks{$^\dag$ Equal contribution. (Corresponding author: Jiankun Wang, Email: wangjk@sustech.edu.cn)}
\thanks{$^{1}$ Shenzhen Key Laboratory of Robotics Perception and Intelligence, Department of Electronic and Electrical Engineering, Southern University of Science and Technology, Shenzhen, China.}%
\thanks{$^{2}$ Research Institute of MA$\&$EI, Peng Cheng Laboratory, Shenzhen, China.}%
\thanks{This work was supported by National Natural Science Foundation of China under Grant 62473191, Guangdong Basic and Applied Basic Research Foundation under Grant 2025A1515012998, and the High level of special funds (G03034K003) from Southern University of Science and Technology, Shenzhen, China.}
}

\begin{document}

\maketitle
\pagestyle{empty}
\thispagestyle{empty}

\begin{abstract}
Service robots have demonstrated significant potential for autonomous trolley collection and redistribution in public spaces like airports or warehouses to improve efficiency and reduce cost. 
Usually, a fully autonomous system for the collection and transportation of multiple trolleys is based on a Leader-Follower formation of mobile manipulators, where reliable docking maneuvers of the mobile base are essential to align trolleys into organized queues. 
However, developing a vision-based robotic docking system faces significant challenges: high precision requirements, environmental disturbances, and inherent robot constraints. 
To address these challenges, we propose a
Disturbance-Aware Visual Predictive Control (DA-VPC) scheme
that incorporates active infrared markers for robust feature extraction across diverse lighting conditions.
This framework explicitly models nonholonomic kinematics and visibility constraints for image-based visual servoing (IBVS), solving the predictive control problem through optimization. It is augmented with an extended state observer (ESO) designed to counteract disturbances during trolley pushing, ensuring precise and stable docking.
Experimental results across diverse environments demonstrate the robustness of this system, with quantitative evaluations confirming high docking accuracy.
\end{abstract}
\section{Introduction}
\begin{figure}[!t]
    \centering
    \includegraphics[width=1\linewidth]{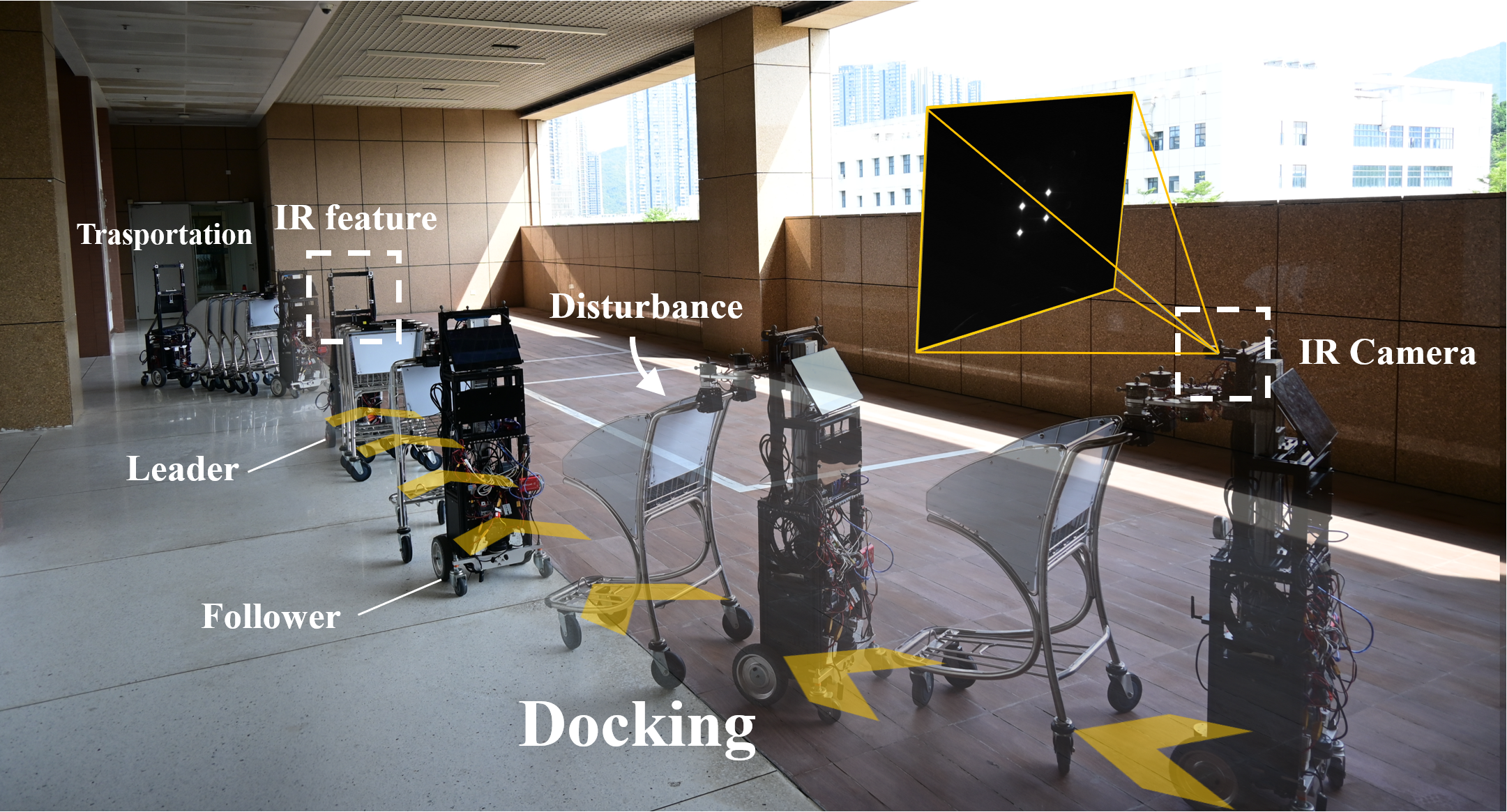}
    \caption{The Follower robot docks the collected trolleys into an organized queue then transport them to designated location. 
    Utilizing visual servoing, it achieves precise docking by observing the infrared features on the Leader robot through its onboard camera.}
    \label{fig:intro}
\vspace{-15pt}
\end{figure}

Mobile manipulation robots are revolutionizing automated transportation by taking over repetitive and heavy-load tasks from humans\cite{Scholz2011cart, tang2024Unwieldy, xie2025Autonomous}.
A promising application is the use of service robots to handle trolleys (i.e., carts), which are deployed in airports, shopping malls, and retail warehouses.
These robots are expected to collect and redistribute trolleys after customer use (e.g., after shopping or airport departures) to improve operational efficiency and reduce costs. 
Specifically, multiple trolleys are required to be collected as a whole unit, and then transported to a designated destination.
Through the cooperation of two robots, this task can be executed in a fully autonomous manner.
As illustrated in Fig. \ref{fig:intro}, one robot pushes a trolley to dock with the other robot, iteratively stacking multiple trolleys into a complete queue.


This paper focuses on the docking maneuvers involved in the sequential stacking of multiple trolleys.
This task imposes strict requirements on both robustness and precision, as misalignment and imperfect connection may lead to collision risks and operational failures.
Our system is built on the dual-arm mobile robot proposed in \cite{zhang2025integrating}.
They developed an adaptive arm controller for pushing trolleys, where mobile manipulation is decoupled into robot arm control and mobile base navigation.
However, a gap remains in target localization and docking control when applying to the trolley stacking task.
Currently, the arm controller regulates pushing forces according to the velocity of the mobile base; therefore, accurate docking maneuvers of the base are essential for adjusting the trolley's pose.
Furthermore, real-world deployments face dynamic environments where pre-built maps become costly to maintain and lack flexibility. LiDAR-based localization also struggles in reflective settings like airports with glass walls and polished floors. In comparison, monocular vision provides a cost-effective solution that directly uses image features to guide docking—offering an elegant and practical solution. 
\begin{figure*}[!t]
    \vspace{10pt}
    \includegraphics[width=1\textwidth]{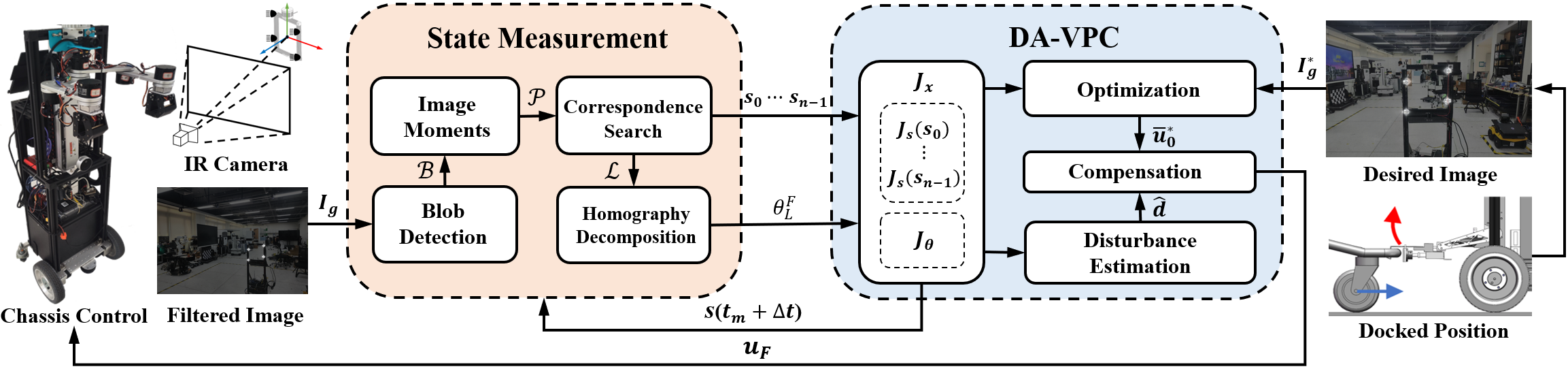}
    \caption{The pipeline for the trolley collection docking process, forming a disturbance-aware visual predictive control framework.}
\label{fig:flow}
\vspace{-10pt}
\end{figure*}

Visual Servoing (VS) offers a promising solution for robotic docking by directly utilizing visual feedback to guide the maneuvers, as highlighted in \cite{li2021survey}.
While VS techniques have been successfully applied to various docking tasks \cite{nie2022Vision, wang2021AutonomousTarget}, their direct implementation in trolley-pushing scenarios remains challenging.
Visual markers for service robots (e.g., AprilTag) are often susceptible to lighting interference and occlusion in real-world environments. Additionally, they require a dedicated, sufficiently large planar surface for installation to ensure visibility for long-range observation.
Existing vision-based docking methods require improvements in robust visual feature extraction and state estimation.
Meanwhile, the system must address a tightly coupled nonlinear control problem under nonholonomic constraints and a limited field-of-view (FOV), necessitating a docking controller designed to ensure precise convergence despite these constraints.
Moreover, such scenarios are subject to significant and non-negligible disturbances—including wheel slippage on uneven ground, inherent kinematic modeling inaccuracies, and persistent perturbations from the pushing mechanism.
These disturbances adversely affect both motion execution and visual observation, leading to servo deviations that undermine docking accuracy and reliability. 
Therefore, effective countermeasures are essential to mitigate external disturbances arising from environmental factors and robot–trolley interactions.

To address the aforementioned challenges in trolley collection, this paper introduces a monocular vision-based docking system that operates without global localization or inter-robot communication. 
For robust perception, an active infrared (IR) marker array on the Leader robot enables consistent visual feature extraction by the Follower robot. 
We develop a holistic kinematic model that incorporates nonholonomic constraints and external disturbances, which serves as the foundation for a visual predictive control framework. 
This framework is augmented with an extended state observer to actively estimate and compensate for unmodeled dynamics. The high precision and robustness of the proposed system are conclusively validated through extensive real-world experiments.
The main contributions of this paper are summarized as follows:
\begin{itemize}
    \item We present a dual-robot system with specialized manipulators for trolley collection, featuring precise docking maneuvers based solely on monocular vision. An active IR marker array is designed as the visual feature, ensuring robust perception under variable lighting conditions. Experiments across diverse real-world environments demonstrate the effectiveness of our system.
    
    \item To handle the disturbances during trolley transportation, an extended state observer is designed to treat the total system disturbance as an additional state for real-time estimation and compensation.

    \item Through kinematic modeling of the image-based visual servoing process for docking, a visual predictive control scheme is introduced to achieve precise maneuvers under nonholonomic constraints and FOV limitations, incorporating disturbance compensation to significantly enhance robustness.
\end{itemize}

    
\section{Related Work}
Recent developments in mobile manipulation have emphasized the importance of human-like pushing skills for expanding payload capacity and service delivery\cite{Scholz2011cart, tang2024Unwieldy}.
For trolley collection, previous studies have investigated trolley identification\cite{xie2025Autonomous}, dual-arm mobile manipulation\cite{zhang2025integrating}, and collaborative queue transportation\cite{xia2023Collaborative}.

Docking is a critical capability for robots, referring to the maneuver of aligning a robot to a target with the desired position and orientation.
It has been widely studied in logistics\cite{Tsiogas2021Pallet, Gutierrez2025Geometric}, quadrotors\cite{nie2022Vision, WANG202240} and surface vehicles\cite{cao2024design}.
Path-tracking methods\cite{herrero2013selfConfiguration, peng2023mpc, yadegar2025AutonomousTarget} rely on high-precision global localization and wireless communication for target pose estimation. 
In contrast, recent studies\cite{wang2021AutonomousTarget, chen2021virtual, nie2022Vision} have leveraged VS to reduce dependency on external modules, thereby developing lightweight systems using cameras. 
For mobile robot docking, Wang et al.\cite{wang2021AutonomousTarget} presented a VS controller using AprilTag for target pose estimation. 
However, the docking problem of trolley stacking faces additional challenges involving model uncertainty and environmental limitations.

A basic VS scheme comprises two essential components: designing visual features associated with the target object and developing a control law to drive the convergence of feature errors.
Visual features are generally defined as geometric primitives (e.g., points and lines).
Recent efforts have explored deep learning-based keypoint detection to improve accuracy and adaptability\cite{zheng2024key}. 
However, learning-based methods rely on training data and can be compromised under poor lighting.
To adapt to diverse environments, recent studies on marker-based techniques propose using an active infrared or ultraviolet LED array, such as a circular fiducial marker\cite{lu2024Fast}.
In addition, Xun et al.\cite{xun2023CREPES} integrated fisheye monocular to enhance relative pose estimation with angle information.
Based on these studies, we develop active IR markers as reliable point features of VS-based docking. 

VS approaches are traditionally classified into Position-based Visual Servoing (PBVS) and Image-based Visual Servoing (IBVS) depending on how they associate the image space with motion space\cite{Chaumette2016}.
PBVS defines control errors by the pose of the camera relative to the target, while IBVS directly controls the movement of visual features to the target frame. 
2.5D VS\cite{malis2121999} combined both methods and extends the error state of the visual features using a pose estimation reference, thereby achieving improved performance. 
Recent advancements involve the integration of VS into Model Predictive Control (MPC) \cite{allibertPredictiveControlConstrained2010, ke2017visual, jin2021gaussian}, forming what is now termed Visual Predictive Control (VPC)—a scheme capable of long-horizon planning while simultaneously enforcing visibility and actuation constraints. 
For nonholonomic robots, Ke et al.\cite{ke2017visual} proposed an approach that achieves convergence by employing a coupled error transformation, which links the robot's orientation to the feature state. 
Jin et al.\cite{jin2021gaussian} addressed unknown actuator dynamics through the online learning of kinematic model with Gaussian process, although the Lorentzian $\rho$-function reduces static error at the expense of motion oscillations.
These approaches rely solely on single-point observations and odometry, without deployment in practical application.

The Extended State Observer (ESO) actively estimates and compensates for the total disturbance—a combination of internal unmodeled dynamics and external disturbances—thereby significantly enhancing the robustness of control systems. Cao et al. \cite{caoESOBasedRobustHighPrecision2024a} employed ESO in an aerial manipulator to estimate dynamic coupling effects between the quadcopter and the robotic arm, enabling high-precision trajectory tracking without relying on a full dynamic model. Tan et al. \cite{tanLESOBasedNMPCTracking2025} combined a Linear ESO with Nonlinear MPC for a climbing robot, rejecting disturbances from slippage and variable-curvature surfaces to achieve high-precision tracking on a wind turbine blade. These studies collectively demonstrate that ESO provides an effective solution for robust and precise control in uncertain environments through active disturbance estimation and compensation.

Inspired by these insights, we propose an IBVS-based approach to address the docking problem in trolley collection by adopting active IR markers as reliable visual features. A VPC framework is employed to handle constraints, while integrated disturbance observation technology compensates for system uncertainties, ultimately achieving precise pose alignment without relying on global localization.

\section{System Overview}
\begin{figure}[!t]
    \centering
    \includegraphics[width=0.8\linewidth]{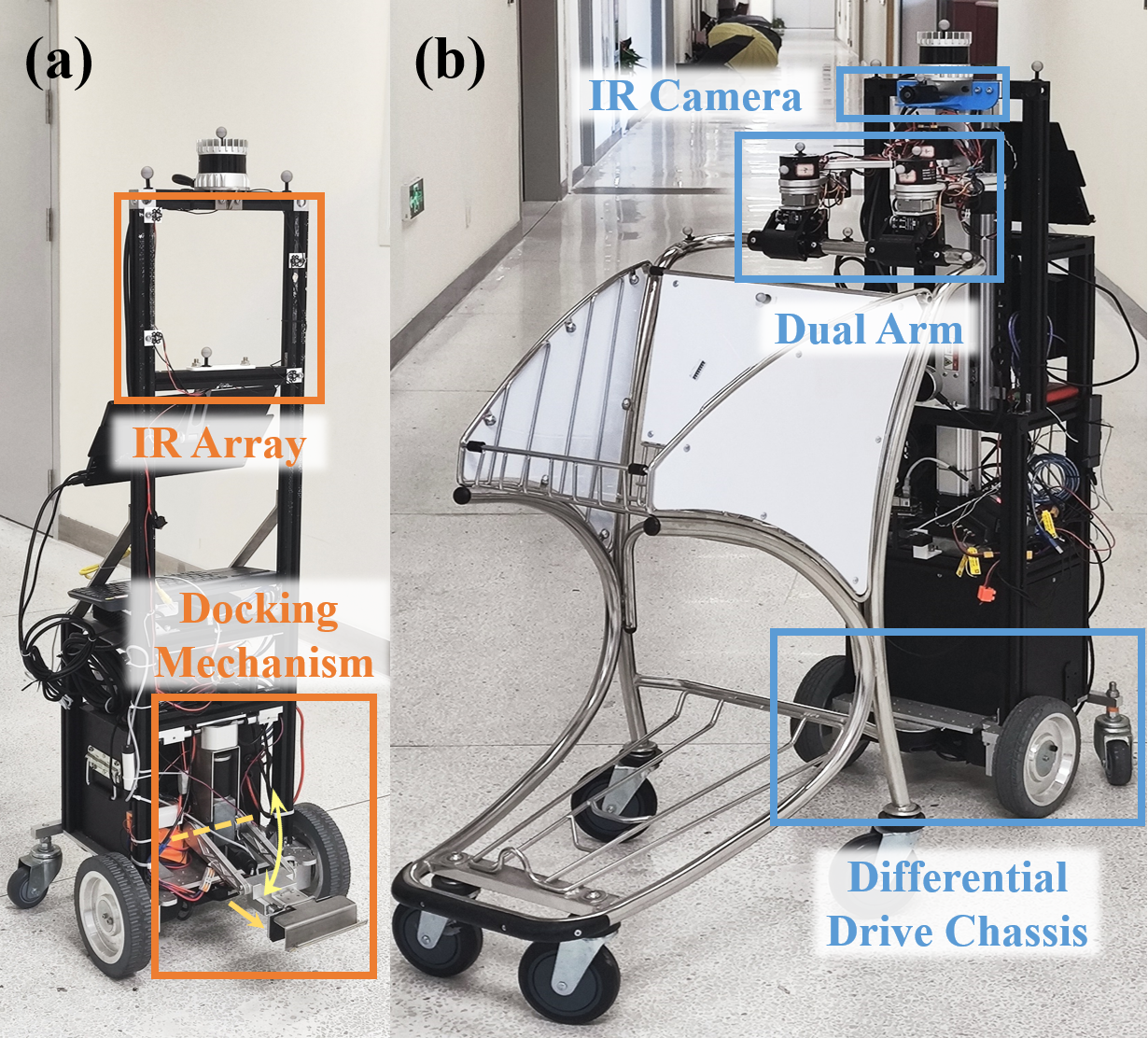} 
    \caption{A prototype of the trolley collection system: (a) Leader robot, (b) Follower robot pushing a trolley.}
    \label{fig:system}
    \vspace{-15pt}
\end{figure}
We introduce a robotic system as a Leader-Follower formation comprising two mobile manipulators for stacking multiple trolleys, which is shown in Fig.~\ref{fig:intro}.
The Follower is in charge of grasping and pushing trolleys, while the Leader, positioned at the head of the trolley queue, serves as the base for docking.
The proposed pipeline that tightly couples with the robotic system is illustrated in Fig.~\ref{fig:flow}.

As shown in Fig.~\ref{fig:system} (a), the Follower utilizes dual Selective Compliance Assembly Robot Arms (SCARA) to stably push the trolley.
Each robotic arm comprises three joint motors and a gripper. Both arms share a coaxial base mounted on a lead screw module for vertical adjustment. 
To effectively manipulate a trolley, the robot typically applies pushing forces along constrained directions by grasping the handle.
Based on the adaptive low-level controller proposed in \cite{zhang2025integrating}, the Follower robot computes the joint torques of the arms to steer the trolley along the moving direction of the chassis.

An active IR LED array (940 nm wavelength) is mounted on the Leader robot as visual markers, as shown in Fig.~\ref{fig:system} (b). 
Correspondingly, a narrowband filter is placed between the camera lens and the CMOS sensor on the Follower robot, which transmits only this specific wavelength of light.
During the docking process, the Follower uses its onboard monocular camera to perceive its state by observing these markers.
The light emitted from each LED marker is projected as an independent, diamond-shaped speck onto the image plane—maintaining a stable shape across distances due to the LED's narrow emission angle—which can be easily extracted as stable visual features.
This hardware design facilitates communication for the robot on a single, human-invisible wavelength, thereby minimizing light pollution in public environments.

Based on the aforementioned vision system, the docking control process can be formulated as follows: the target state is defined by a reference image captured at the docked position, which serves as the desired image for visual servoing. The docking controller computes the Follower's chassis velocity by driving the current image toward the desired one.
This scheme consists of two components: state measurement and DA-VPC.
The system processes the current image from the Follower's camera by performing threshold segmentation to extract the LED spots and computing the centroid of image moments, allowing for the resolution of relative rotation through homography decomposition.
This enables an extended system state equation for image-based docking control, whereby the nominal system is solved through optimization, while an ESO estimates and compensates for disturbances in real time. This integrated approach generates constrained velocity commands to ensure robust convergence and utilizes predicted feature motion for future correspondence matching. Further methodological details will be provided in subsequent sections.

The Follower determines the termination of the docking by the joint torque variations of its arm and the visual servoing error.
Upon reaching the docking position, the Follower steers the trolley forward until all trolleys in the queue are rigidly connected. 
Notably, for docking with the first trolley, a gravity-based self-locking guide structure is designed as the end effector of the Leader. 
When the trolley’s front beam contacts this guide structure, the front beam exerts a force along the direction of motion. In response, the guide structure generates a perpendicular reaction force (normal to the ground), which passively lifts the end effector. 
As the trolley moves further, the end effector first descends under gravitational force, then a linear actuator engages to securely lock the trolley's front beam.
With the docking process completed, the Follower retracts its grippers and resumes its collection mission.
\section{Problem Formulation}

\subsection{Notations}  

\begin{figure}[!h]
    \vspace{-0.3cm}
    \centering
    \includegraphics[width=1\linewidth]{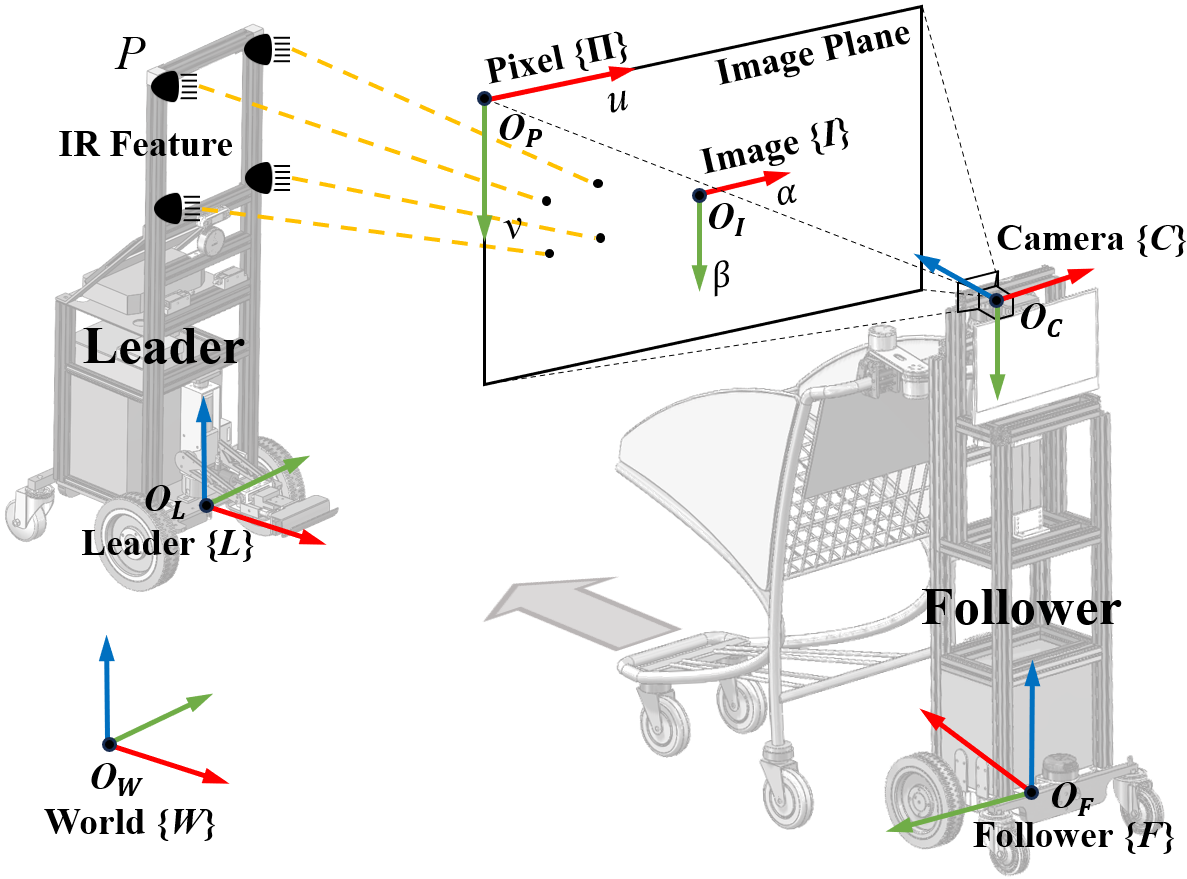} 
    \caption{Coordinate system description for robot docking process.}
    \label{fig:frame}
    \vspace{-5pt}
\end{figure}

The mathematical notation in this work follows these conventions: vectors and matrices are represented in boldface (e.g., $\boldsymbol{s}$ for vectors and $\boldsymbol{K}$ for matrices), while sets are denoted using calligraphic font (e.g., $\mathcal{B}$). 
The transpose operation is indicated by $^\top$ (e.g., $\boldsymbol{s}^\top$). 
Additionally, superscripts are employed to specify reference frames, such as $x^F_L$ representing the x-coordinate position of the Leader robot in frame $\{F\}$.
The coordinate frames employed in the formulation are enumerated below and illustrated in Fig.~\ref{fig:frame}:

1) Follower Frame $\{F\}$: Fixed on the chassis of Follower robot, velocity commands are expressed in this frame.

2) Leader Frame $\{L\}$: The spatial configuration of the visual markers is represented in this frame.

3) Camera Frame $\{C\}$: Mounted at the camera's optical center, with its z-axis coincident with the optical axis.

4) Pixel Frame $\{\Pi\}$: With its origin at the image's top-left corner, marker projections are captured in this frame.

5) Image Frame $\{I\}$: Normalized image plane coordinates, which facilitate the derivation of the state equation.
\subsection{Kinematic Model for Visual Servoing Docking}
The differentially driven Follower robot follows the unicycle model with control input $\boldsymbol{u_F} = [v_F, \omega_F]^\top$, consisting of linear velocity and angular velocity:
\begin{equation}
    \begin{cases} 
    \dot{x}_F = v_F \cos \theta_F \\
    \dot{y}_F = v_F \sin \theta_F \\
    \dot{\theta}_F = \omega_F,
    \end{cases}
\label{eq:unicycle}
\end{equation}
where $[x_F, y_F, \theta_F]^\top$ denotes the Cartesian coordinates and the heading angle of the robot on $XOY$ plane.
Differentiation yields the relative velocity of the Leader robot \(\{L\}\) with respect to the Follower frame \(\{F\}\) as:
\begin{equation}
    \begin{cases} 
        \dot{x}_L^F = y_L^F \omega_F - v_F \\
        \dot{y}_L^F = -x_L^F \omega_F \\
        \dot{\theta}_L^F = -\omega_F. 
    \end{cases}
    \label{eq:relative}
\end{equation}
For a feature point $P$ on the Leader robot, its coordinate in the Leader frame $\{L\}$ is $\boldsymbol{P}^L=[X_P^L, Y_P^L, Z_P^L]^\top$. The coordinate of this point expressed in the Follower frame \(\{F\}\) is:
\begin{equation}
\begin{cases}
X_P^F = x_L^F + X_P^L \cos\theta_L^F - Y_P^L \sin\theta_L^F \\
Y_P^F = y_L^F + X_P^L \sin\theta_L^F + Y_P^L \cos\theta_L^F \\
Z_P^F = Z_P^L.
\end{cases}
\label{eq:point_in_F}
\end{equation}
The camera's coordinate under the Follower frame \(\{F\}\) is $\boldsymbol{\xi} = [X_C^F, Y_C^F, Z_C^F]^\top$, through transformation between the camera frame \(\{C\}\) and \(\{F\}\), this feature point $\boldsymbol{P}^C=[X_P^C,Y_P^C,Z_P^C]^\top$ in frame \(\{C\}\) can be expressed as:
\begin{equation}
\begin{cases}
X_P^C = -y_L^F + Y_C^F - X_P^L \sin \theta_L^F - Y_P^L \cos \theta_L^F \\
Y_P^C = Z_C^F - Z_P^L \\
Z_P^C = x_L^F - X_C^F + X_P^L \cos \theta_L^F - Y_P^L \sin \theta_L^F.
\end{cases}
\label{eq:featurepoint}
\end{equation}
By differentiating \eqref{eq:featurepoint} and substituting \eqref{eq:relative} into equation, we obtain the motion of feature point within the camera:
\begin{equation}
\begin{cases}
\begin{aligned}
    \dot{X_P^C} &= -\dot{y}_L^F-\omega_L^F(X_P^L\cos\theta_L^F-Y_P^L\sin\theta_L^F) \\
                &= (Z_P^C + X_C^F)\omega_F \\
    \dot{Z_P^C} &= \dot{x}_L^F-\omega_L^F(X_P^L\sin\theta_L^F+Y_P^L\cos\theta_L^F) \\
                &= -v_F - (X_P^C - Y_C^F) \omega_F.
\end{aligned}
\end{cases}
\label{eq:featurepointmotion}
\end{equation}

Based on the pinhole camera model, the projection relationship between a homogeneous pixel coordinate $\boldsymbol{p_h} = [u,v,1]^\top$ in frame $\{\Pi\}$ and the feature point in frame $\{C\}$ is given by:
\begin{equation}
Z_P^C\boldsymbol{p_h} = \boldsymbol{K}\boldsymbol{P}^C,
\label{eq:pinhole_projection}
\end{equation}
where $\boldsymbol{K} \in \mathbb{R}^{3\times3}$ is the camera intrinsic matrix. We use $\boldsymbol{p}=[u,v]^\top$ to denote the coordinates in frame $\{\Pi\}$ and $\boldsymbol{s} = [\alpha,\beta]^\top$ to denote the coordinates in frame $\{I\}$. Using the pinhole model in \eqref{eq:pinhole_projection}, the image coordinates can be expressed as:
\begin{equation}
\alpha = \frac{X_P^C}{Z_P^C},\quad \beta = \frac{Y_P^C}{Z_P^C} = \frac{Z_C^F - Z_P^L}{Z_P^C}.
\label{eq:depthscale}
\end{equation}
By taking the time derivative of $\boldsymbol{s}$, we can obtain:
\begin{equation}
\dot{\boldsymbol{s}} = \begin{bmatrix}
\dot{\alpha} \\ \dot{\beta}
\end{bmatrix}
 = 
\begin{bmatrix}
(\dot{X_P^C} - \alpha\dot{Z_P^C})/Z_P^C \\ (\dot{Y_P^C} - \beta\dot{Z_P^C})/Z_P^C
\end{bmatrix}.
\label{eq:normalized_deriv}
\end{equation}
With the known camera-feature height difference defined as $\Delta h = Z_C^F - Z_P^L \neq 0$ and the camera installation offset as $[X_C^F, Y_C^F]^\top=[t_x,t_y]^\top$, the substitution of \eqref{eq:featurepointmotion} and \eqref{eq:depthscale} into \eqref{eq:normalized_deriv} leads to:
\begin{equation}
\dot{\boldsymbol{s}} = \boldsymbol{J_s(s)} \boldsymbol{u_F},
\end{equation}
where:
\begin{equation}
\boldsymbol{J_s(s)} =
\begin{bmatrix}
\frac{\alpha\beta}{\Delta h} & 1 + \alpha^2 + \frac{\beta(t_x - \alpha t_y)}{\Delta h} \\
\frac{\beta^2}{\Delta h} & \alpha\beta - \frac{\beta^2 t_y}{\Delta h}
\end{bmatrix}.
\label{eq:interaction_matrix}
\end{equation}

This matrix, known as the image Jacobian in IBVS, provides a mapping from robot motion to image point motion, forming the basis for vision-based control that minimizes feature errors. 
However, defining the system state purely in terms of feature coordinates $\boldsymbol{s}$ may lead to local minima and steady-state errors due to the nonholonomic constraints of the differentially-driven chassis. 
Motivated by 2.5D VS in \cite{malis2121999}, we address this limitation by formulating the system state $\boldsymbol{x}$ as a combination of image coordinates and relative orientation between robots: $\boldsymbol{x}=[\boldsymbol{s}_0^\top, \dots, \boldsymbol{s}_{n-1}^\top, \theta_L^F]^\top\in\mathbb{R}^{2n+1}$, where $n$ is the number of feature points. Correspondingly, the desired state is defined as $\boldsymbol{x}^*=[\boldsymbol{s}_0^{*\top}, \dots, \boldsymbol{s}_{n-1}^{*\top}, 0]^\top$, where $\boldsymbol{s}^*_{\cdot}$ denotes the feature coordinates at the desired docking pose and the zero orientation element ensuring proper alignment. The state error is obtained:
$\boldsymbol{e}(t) = \boldsymbol{x}(t) - \boldsymbol{x}^*$, the docking control objective is to design $\boldsymbol{u_F}(t)$ to drive the Follower to the desired docking state:
\begin{equation}
    \lim_{t\to\infty} ||\boldsymbol{e}(t)|| = \boldsymbol{0}.
\end{equation}

During this transportation process, the weight of the trolley, which acts as a variable load and induces noticeable speed fluctuations in the Follower robot, thereby significantly influencing its motion dynamics.
Furthermore, complex terrain conditions, such as uneven surfaces and variations in ground friction, introduce external disturbances that further compromise the stability and tracking accuracy of the visual servo system.
To analyze the effect of these disturbances on system performance, the system model is augmented with disturbance terms, leading to the following unified kinematics formulation:
\begin{equation}
\dot{\boldsymbol{x}} = \boldsymbol{J_x(x)} \boldsymbol{u_F} + \boldsymbol{d},
\label{eq:disturbed_dynamics}
\end{equation}
where $\boldsymbol{J_x(x)} \in \mathbb{R}^{(2n+1) \times 2}$ denotes the extended image Jacobian matrix that accounts for both the image feature dynamics and the relative orientation:
\begin{equation}
\boldsymbol{J_x(x)} = \left[ \boldsymbol{J_s}(\boldsymbol{s}_0)^\top, \dots, \boldsymbol{J_s}(\boldsymbol{s}_{n-1})^\top, \boldsymbol{J_{\theta}}^\top \right]^\top,
\end{equation}
with $\boldsymbol{J_{\theta}} = [0, -1]$. 
The lumped disturbance vector $\boldsymbol{d} = [\boldsymbol{d_s}^\top, d_\theta]^\top \in \mathbb{R}^{2n+1}$ includes disturbances in image feature dynamics $\boldsymbol{d_s}\in\mathbb{R}^{2n}$ from load variations and terrain-induced vibrations, as well as orientation-related disturbances $d_\theta \in \mathbb{R}$ from wheel slippage and estimation errors.
\section{Disturbance-Aware Visual Predictive Docking Control}
\subsection{Homography Based State Measurement}
For each captured image frame, a threshold $\lambda_I$ is applied to the acquired grayscale image $\boldsymbol{I_g}(u, v)$ to facilitate subsequent processing:
\begin{equation}
\boldsymbol{I_g^{\prime}}(u, v) = \begin{cases} 
\boldsymbol{I_g}(u, v), & \text{if } \boldsymbol{I_g}(u, v) > \lambda_I \\ 
0, & \text{otherwise}.
\end{cases}
\label{eq:grayscale}
\end{equation}
We perform blob detection and segmentation on $\boldsymbol{I_g^{\prime}}(u, v)$ and obtain a set of detected blobs $\mathcal{B} = \{\boldsymbol{b}_0, \dots, \boldsymbol{b}_{N-1}\}$, where each $\boldsymbol{b}_i$ represents a connected component in the thresholded image. We compute the centroid of each $\boldsymbol{b}_i$ with subpixel-level precision using image moments:
\begin{equation}
M_{pq}^{i} = \sum_{(u, v) \in \boldsymbol{b}_i} u^p v^q \boldsymbol{I_g^{\prime}}(u, v), 
\label{eq:moment_defination}
\end{equation}
\begin{equation}
\boldsymbol{p}_i = [\hat{u_i}, \hat{v_i}]^\top = [{M_{10}^{i}}/{M_{00}^{i}}, {M_{01}^{i}}/{M_{00}^{i}}]^\top,
\label{eq:moment_point}
\end{equation}
where $p+q$ is the order of the image moments. This results in a set of candidate image points $\mathcal{P} = \{\boldsymbol{p}_0, \dots, \boldsymbol{p}_{N-1}\}$ representing the centroids of the detected blobs.
In complex lighting environments, interference may occur due to other devices emitting light in the same wavelength band. 
To identify valid IR feature projections among the detected points, we extract the feature set $\mathcal{L} =\{\boldsymbol{p}_0, \dots, \boldsymbol{p}_{n-1}\} \subseteq \mathcal{P}$ by leveraging the spatial configuration and geometric constraints of the IR feature, while simultaneously recovering their correspondence. 

Since the combinatorial point-matching method incurs high computational complexity when dealing with a large number of points, we instead establish a relationship between camera motion and the resulting optical flow of feature points. Based on \eqref{eq:interaction_matrix}, we predict the future feature coordinates at time $t_m+\Delta t_m$ from the measurement at time $t_m$:
\begin{equation}
\boldsymbol{s}(t_m+\Delta t_m) = \boldsymbol{s}(t_m) + \boldsymbol{J_s}(\boldsymbol{s}(t_m))\boldsymbol{u_F}(t_m)\Delta t_m.
\label{eq:feature_prediction}
\end{equation}
Each detected feature point inherits the previous matching result if its distance to the nearest predicted point falls below a predefined threshold.

With the feature set $\mathcal{L}$ established, the image coordinates $\boldsymbol{s}_0, \dots, \boldsymbol{s}_{n-1}$ and the corresponding target coordinates $\boldsymbol{s}_0^*, \dots, \boldsymbol{s}_{n-1}^*$ from the desired image $\boldsymbol{I_g}^*$ are computed via \eqref{eq:pinhole_projection}. 
By leveraging the coplanarity of these points, a homography between the two camera poses can be estimated. The state $\theta_L^F$ is determined by the relative rotation about the camera's y-axis, which is derived through homography decomposition \cite{MotionandStructure}.

\subsection{Disturbance Observer Design}

An ESO is introduced to treat the total system disturbance as an additional state for real-time estimation and compensation, without requiring precise prior knowledge of the disturbance dynamics:
\begin{equation}
\begin{cases}
\dot{\boldsymbol{x}} = \boldsymbol{J_x}(\boldsymbol{x})\boldsymbol{u_F} + \boldsymbol{d} \\
\dot{\boldsymbol{d}} = \boldsymbol{h}(t),
\end{cases}
\end{equation}
where $\boldsymbol{h}(t) \in \mathbb{R}^{2n+1}$ represents the unknown but bounded time derivative of the disturbance. The observer is formulated as:
\begin{equation}
\begin{cases}
\dot{\hat{\boldsymbol{x}}} = \boldsymbol{J_x}(\boldsymbol{x})\boldsymbol{u_F} + \hat{\boldsymbol{d}} + \boldsymbol{L}_1 (\boldsymbol{y} - \hat{\boldsymbol{x}}) \\
\dot{\hat{\boldsymbol{d}}} = \boldsymbol{L}_2 (\boldsymbol{y} - \hat{\boldsymbol{x}}),
\end{cases}
\label{eq:eso}
\end{equation}
where $\hat{\boldsymbol{x}}$ and $\hat{\boldsymbol{d}}$ are the estimates of the system state and disturbance, respectively. $\boldsymbol{y}$ is the measured system output, assumed to be the full state in system. $\boldsymbol{L}_1$ and $\boldsymbol{L}_2$ are observer gain matrices to be designed.

Define the observation errors as $\tilde{\boldsymbol{x}} = \boldsymbol{x} - \hat{\boldsymbol{x}}$ and $\tilde{\boldsymbol{d}} = \boldsymbol{d} - \hat{\boldsymbol{d}}$, the error dynamics with $\boldsymbol{\epsilon} = [\tilde{\boldsymbol{x}}^\top, \tilde{\boldsymbol{d}}^\top]^\top$ are given by:
\begin{equation}
\dot{\boldsymbol{\epsilon}} = \boldsymbol{A} \boldsymbol{\epsilon} + \boldsymbol{D} \boldsymbol{h}(t),
\label{eq:error_dynamics}
\end{equation}
where:
\begin{equation}
\boldsymbol{A} = \begin{bmatrix} -\boldsymbol{L}_1 & \boldsymbol{I} \\ -\boldsymbol{L}_2 & \boldsymbol{0} \end{bmatrix}, \quad
\boldsymbol{D} = \begin{bmatrix} \boldsymbol{0} \\ \boldsymbol{I} \end{bmatrix}.
\label{eq:error_matrix}
\end{equation}
The matrix $\boldsymbol{A}$ can be made Hurwitz through proper design of $\boldsymbol{L}_1$ and $\boldsymbol{L}_2$, the explicit solution of \eqref{eq:error_dynamics} is given by:
\begin{equation}
\boldsymbol{\epsilon}(t) = e^{\boldsymbol{A} t} \boldsymbol{\epsilon}(0) + \int_0^t e^{\boldsymbol{A} (t-\tau)} \boldsymbol{D} \boldsymbol{h}(\tau) d\tau.
\end{equation}
Consider the $i$-th component of the error vector $\boldsymbol{\epsilon}(t)$:
\begin{equation}
\begin{aligned}
|\epsilon_i(t)| &= \left| \left[ e^{\boldsymbol{A} t} \boldsymbol{\epsilon}(0) + \int_0^t e^{\boldsymbol{A} (t-\tau)} \boldsymbol{D} \boldsymbol{h}(\tau) d\tau \right]_i \right| \\
&\leq \underbrace{\left| \left[ e^{\boldsymbol{A} t} \boldsymbol{\epsilon}(0) \right]_i \right|}_{E_1} + \underbrace{\left| \left[ \int_0^t e^{\boldsymbol{A} (t-\tau)} \boldsymbol{D} \boldsymbol{h}(\tau) d\tau \right]_i \right|}_{E_2}.
\end{aligned}
\end{equation}
Since $\boldsymbol{A}$ is Hurwitz, there exists a constant $\mu > 0$ such that $|(e^{\boldsymbol{A} t})_{ij}| \leq \mu$ for all $i, j$ and $t \geq 0$. Then:
\begin{equation}
E_1 = \left| \sum_{j=1}^{4n+2} (e^{\boldsymbol{A} t})_{ij} \epsilon_j(0) \right| \leq \sum_{j=1}^{4n+2} |(e^{\boldsymbol{A} t})_{ij}| \cdot |\epsilon_j(0)| \leq \mu \epsilon_a(0).
\end{equation}
For $E_2$, using the matrix calculus identity:
\begin{equation}
\int_0^t e^{\boldsymbol{A} (t-\tau)} \boldsymbol{D} d\tau = \boldsymbol{A}^{-1} e^{\boldsymbol{A} t} \boldsymbol{D} - \boldsymbol{A}^{-1} \boldsymbol{D}.
\end{equation}
Under the assumption that $\boldsymbol{h}(t)$ is bounded such that $\gamma \triangleq \sup_{t\geq0} \|\boldsymbol{h}(t)\|<\infty$, the upper bound of $E_2$ can be derived as:
\begin{equation}
E_2 \leq \gamma \left| \left[\boldsymbol{A}^{-1} e^{\boldsymbol{A} t} \boldsymbol{D} \right]_i \right| + \gamma \left| \left[ \boldsymbol{A}^{-1} \boldsymbol{D} \right]_i \right| \leq \gamma (\kappa + \sigma),
\end{equation}
where $\kappa=\sup_{t \geq 0}\max_i |[\boldsymbol{A}^{-1} e^{\boldsymbol{A} t} \boldsymbol{D}]_{i}|$ and $\sigma=\max_i |[\boldsymbol{A}^{-1} \boldsymbol{D}]_{i}|$.

Combining $E_1$ and $E_2$ yields the ultimate error bound:
$|\epsilon_i(t)| \leq \mu \epsilon_a(0) + \gamma (\kappa + \sigma)$. 
In particular, when initial error is zero $(\epsilon_a(0) = 0)$:
$|\epsilon_i(t)| \leq \gamma (\kappa + \sigma)$. 
This proves that the observation error $\boldsymbol{\epsilon}(t)$ is uniformly ultimately bounded, and the error bound can be reduced by increasing the observer gains (which decreases $\kappa$ and $\sigma$ through larger $\boldsymbol{L}_1$, $\boldsymbol{L}_2$), though practical implementation requires balancing estimation accuracy against noise amplification sensitivity.

\subsection{Disturbance-Aware Visual Predictive Control}
By deriving the visual servoing prediction model without considering disturbance effects and applying forward Euler discretization with the control period $\Delta t_c$, we obtain the nominal discrete-time prediction model as:
\begin{equation}
\label{eq:dis_dynamics}
\boldsymbol{\bar{x}}_{k+1} = f(\boldsymbol{\bar{x}}_k, \boldsymbol{\bar{u}}_k) =\boldsymbol{\bar{x}}_k + \Delta t_c\boldsymbol{J_x}(\boldsymbol{\bar{x}}_k)\boldsymbol{\bar{u}}_k,
\end{equation}
where $\boldsymbol{\bar{x}}_k$ and $\boldsymbol{\bar{u}}_k$ represent the nominal system state and nominal control input at time step $k$, respectively.

The visual predictive controller optimizes a sequence of control inputs $\{\boldsymbol{\bar{u}}_0, \dots, \boldsymbol{\bar{u}}_{N_p-1}\}$ of Follower robot over a prediction horizon $N_p$ by minimizing a cost function:
\begin{subequations}
\begin{align}
\min_{\boldsymbol{\bar{u}}_0, \dots, \boldsymbol{\bar{u}}_{N_p-1}} &
\left\| \boldsymbol{\bar{x}}_{N_p} - \boldsymbol{x}^* \right\|_{\boldsymbol{P}}^2 + \sum_{k=0}^{N_p-1} \left\| \boldsymbol{\bar{u}}_k \right\|_{\boldsymbol{R}}^2\\
\text{s.t.} \quad
& \boldsymbol{\bar{x}}_{k+1} = f(\boldsymbol{\bar{x}}_k, \boldsymbol{\bar{u}}_k)\\
& \boldsymbol{\bar{x}}_0 = \boldsymbol{x}_{init}\\
& \boldsymbol{s}_{\min} \leq \boldsymbol{\bar{s}}_k \leq \boldsymbol{s}_{\max}\\
& \boldsymbol{u}_{\min} \leq \boldsymbol{\bar{u}}_k \leq \boldsymbol{u}_{\max}\\
& \Delta \boldsymbol{u}_{\min} \leq \boldsymbol{\bar{u}}_{k+1}- \boldsymbol{\bar{u}}_{k} \leq \Delta \boldsymbol{u}_{\max},
\end{align}
\label{eq:nmpc_problem}
\end{subequations}
where $\boldsymbol{x}^*$ is the desired terminal state; $\boldsymbol{P}, \boldsymbol{R}$ are positive definite weighting matrices for terminal state error and control effort; $\boldsymbol{s}_{\min}, \boldsymbol{s}_{\max}$ are the visibility constraints of FOV; $\boldsymbol{u}_{\min}, \boldsymbol{u}_{\max}$ are the lower and upper bounds for velocity; $\Delta \boldsymbol{u}_{\min}, \Delta \boldsymbol{u}_{\max}$ are the lower and upper bounds for velocity variation.

At each control instant $t_c$, the system measures the current state $\boldsymbol{x}(t_c)$, utilizes the disturbance observer in \eqref{eq:eso} to estimate the disturbance $\hat{\boldsymbol{d}}(t_c)$, and solves the nominal visual predictive control problem \eqref{eq:nmpc_problem} to obtain the nominal control sequence $\{\boldsymbol{\bar{u}}_0^*, \dots, \boldsymbol{\bar{u}}_{N_p-1}^*\}$. The actual control input is then computed by augmenting the nominal control with a disturbance compensation term.
Considering the disturbed system dynamics given by \eqref{eq:disturbed_dynamics}, the kinematics at time $t_c$ can be expressed as:
\begin{equation}
\dot{\boldsymbol{x}}(t_c) = \boldsymbol{J_x}(\boldsymbol{x}(t_c))(\boldsymbol{\bar{u}}_0^* + \boldsymbol{u_C}) + \hat{\boldsymbol{d}}(t_c).
\label{eq:compensation_dynamics}
\end{equation}
To compensate for the disturbance, we design $\boldsymbol{u_C}$ such that it best cancels out $\hat{\boldsymbol{d}}(t_c)$ in a least-squares sense, by solving:
\begin{equation}
\min_{\boldsymbol{u_C}} \left\| \boldsymbol{J_x}(\boldsymbol{x}(t_c))\boldsymbol{u_C} + \hat{\boldsymbol{d}}(t_c) \right\|^2,
\label{eq:minsquare}
\end{equation}
which leads to the final control law with disturbance compensation:
\begin{equation}
\boldsymbol{u_F}(t_c) = \text{sat}\left( \boldsymbol{\bar{u}}_0^* - \boldsymbol{J_x^\dag}(\boldsymbol{x}(t_c)) \hat{\boldsymbol{d}}(t_c),\ \boldsymbol{u}_{\min},\ \boldsymbol{u}_{\max} \right),
\label{eq:getuf}
\end{equation}
where $\text{sat}(\cdot)$ denotes the element-wise saturation function and $\boldsymbol{J_x^\dag}$ denotes the Moore–Penrose pseudoinverse of $\boldsymbol{J_x}$.
This composite strategy is designed to achieve nominal performance through $\boldsymbol{\bar{u}}_0^*$ while actively rejecting disturbances via feedforward compensation, with the saturation function ensuring that the total control input respects the system's physical limits $\boldsymbol{u}_{\min}$ and $\boldsymbol{u}_{\max}$. The whole workflow of the proposed DA-VPC is illustrated in Algorithm \ref{alg:da_vpc}.
\begin{algorithm}[!t]
\caption{Disturbance Aware Visual Predictive Control (DA-VPC)}
\label{alg:da_vpc}
\begin{algorithmic}[1]
\STATE\textbf{Input}: Current image $\boldsymbol{I_g}$, desired image $\boldsymbol{I_g}^*$, camera installation position $\boldsymbol{\xi}$, feature installation position $\boldsymbol{P}^L$, tolerance $\boldsymbol{\varepsilon}$.\\
\STATE\textbf{Initialization}: Initialize system state $\boldsymbol{x}(0)$, disturbance observer states $\hat{\boldsymbol{x}}(0)$ and $\hat{\boldsymbol{d}}(0)$, compute $\boldsymbol{x}^*$ from $\boldsymbol{I_g}^*$.
\WHILE{$|| \boldsymbol{x}-\boldsymbol{x}^*|| > \boldsymbol{\varepsilon}$}
    \STATE Obtain candidate image point set $\mathcal{P}$ through \eqref{eq:grayscale}, \eqref{eq:moment_defination} and \eqref{eq:moment_point}.
    \STATE Extract the feature set $\mathcal{L}$ via feature spatial configuration, and compute $\boldsymbol{s}_0, \dots, \boldsymbol{s}_{n-1}$ using \eqref{eq:pinhole_projection}.
    \STATE Compute $\theta_L^F$ using homography decomposition.
    \STATE Estimate disturbance $\hat{\boldsymbol{d}}$ using \eqref{eq:eso}.
    \STATE Solve nominal VPC problem \eqref{eq:nmpc_problem} for optimal control sequence $\{\boldsymbol{\bar{u}}_0^*, \dots, \boldsymbol{\bar{u}}_{N_p-1}^*\}$.
    \STATE Compute disturbance compensation $\boldsymbol{u_C}$ using \eqref{eq:minsquare}.
    \STATE Calculate and apply Follower robot control input $\boldsymbol{u_F}$ by \eqref{eq:getuf}.
    \STATE Update the feature prediction by \eqref{eq:feature_prediction}.
\ENDWHILE
\end{algorithmic}
\end{algorithm}
\section{Experiments}

\subsection{Experimental Setup}
Comprehensive experimental validation of the proposed docking control method is presented through both simulation and real-world implementations. 
We design the marker array by 4 high-power IR LEDs (3W output, 60$^\circ$ beam angle) with a constant-current driver, maintaining stable illumination.
For visual perception, we employ a MV-SUA133GM industrial camera with 120$^\circ$ FOV providing 1280×1024 resolution. 
Due to the addition of the filter in camera, visible light can hardly pass through. Therefore, we employ a 940 nm light source to illuminate the calibration board, enabling convenient and efficient calibration.
Each robot is equipped with an Intel NUC (Core i7-1165G7 CPU@4.70GHz, 32GB RAM) for onboard computation. 
The software system is deployed on Ubuntu 20.04 with ROS2 Galactic, where the control problem is formulated and solved using CasADi\cite{Andersson2019} in real time. 
Through node synchronization, this implementation maintains a 60 Hz perception frequency while achieving a 20 Hz control frequency.

The docking performance is quantitatively evaluated using the following metrics:

1) $\boldsymbol{E_{p}}$ - Position error between the terminal and the desired pose of the Follower robot, composed of the normal docking error $e_x$ and lateral docking error $e_y$, with $e_x = |x^{L+}_{F}-x^{L*}_{F}|$, $e_y = |y^{L+}_{F}-y^{L*}_{F}|$, where $[x^{L+}_{F},y^{L+}_{F}]^\top$ and $[x^{L*}_{F},y^{L*}_{F}]^\top$ denote the terminal and desired position in Leader frame $\{L\}$, respectively.

2) $\boldsymbol{E_{s}}$ - State error between the terminal and desired system states, including $e_\theta$ and $e_p$. Here, $e_{\theta}=|\theta^{F+}_{L}-\theta^{F*}_{L}|$ represents the angular deviation between the terminal and desired states. $e_p$ denotes the Root Mean Square (RMS) distance error of $n$ feature points in the pixel frame $\{\Pi\}$, calculated as: $e_p = \sqrt{\frac{1}{n} \|\boldsymbol{p}^+ - \boldsymbol{p}^*\|^2}$, where $\boldsymbol{p}^+$ and $\boldsymbol{p}^*$ are the terminal and desired coordinate vectors, respectively.

3) $\boldsymbol{M_{sm}}$ - Smoothness of the docking motion, calculated by the RMS of the velocity variation $\Delta\boldsymbol{u}=[\Delta v, \Delta w]^\top$ in each trial: $\boldsymbol{M_{sm}} = \sqrt{\frac{1}{T_{task}}\sum_{t=0}^{T_{task}}{\Delta\boldsymbol{u}}^2}$, where $T_{task}$ denotes the task execution time for docking control. 

\begin{figure}[!t]
    \centering
    \includegraphics[width=1\linewidth]{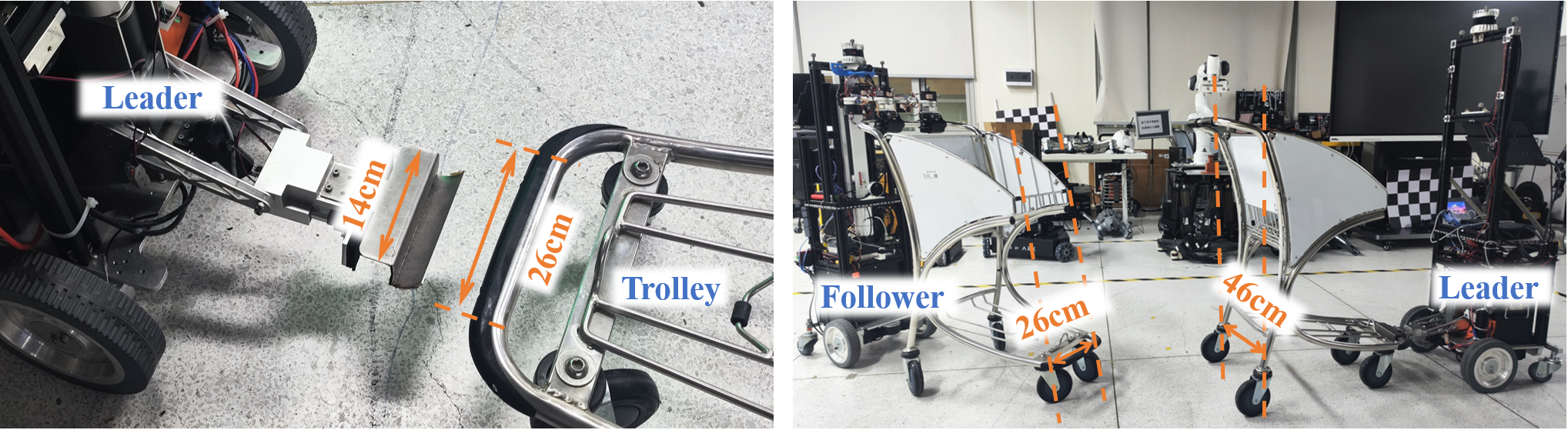}
    \caption{Hardware platform of experiments, with key dimensions illustrated.}
    \label{fig:target}
    \vspace{-15pt}
\end{figure}

As shown in Fig.~\ref{fig:target}, the key dimensions of the robot's actuator and the trolley impose a requirement on the docking accuracy. Specifically, the lateral docking error is constrained by two different upper bounds: for the first trolley, which docks directly to the Leader robot’s actuator, the lateral error $e_y^{(1)}$ satisfies $e_y^{(1)} \le \bar{e}_y^{(1)} = 6~\text{cm}$, whereas for each subsequent trolley, docking to the preceding one, the lateral error $e_y^{(k)}$ satisfies $e_y^{(k)} \le \bar{e}_y^{(k)} = 10~\text{cm}$.


\subsection{Simulation Experiments}
To evaluate the docking performance of the proposed nominal kinematic model and controller, we conducted simulation experiments and compared its performance with several IBVS methods, including the basic IBVS approach \cite{Chaumette2016} and the method by Ke et al. \cite{ke2017visual}. The Leader robot remained stationary at $[0\,\text{m}, 0\,\text{m}]$, while the desired docking state for the Follower robot was $[1.5\,\text{m}, 0\,\text{m}, 0^\circ]$. Simulation parameters precisely matched the physical hardware specifications, including camera intrinsics and the installation poses of the camera and features.

\begin{figure}[!t]
\vspace{10pt}
    \subfloat[Docking trajectories in XOY Plane\label{fig:curve_a}]{
        \includegraphics[width=\linewidth]{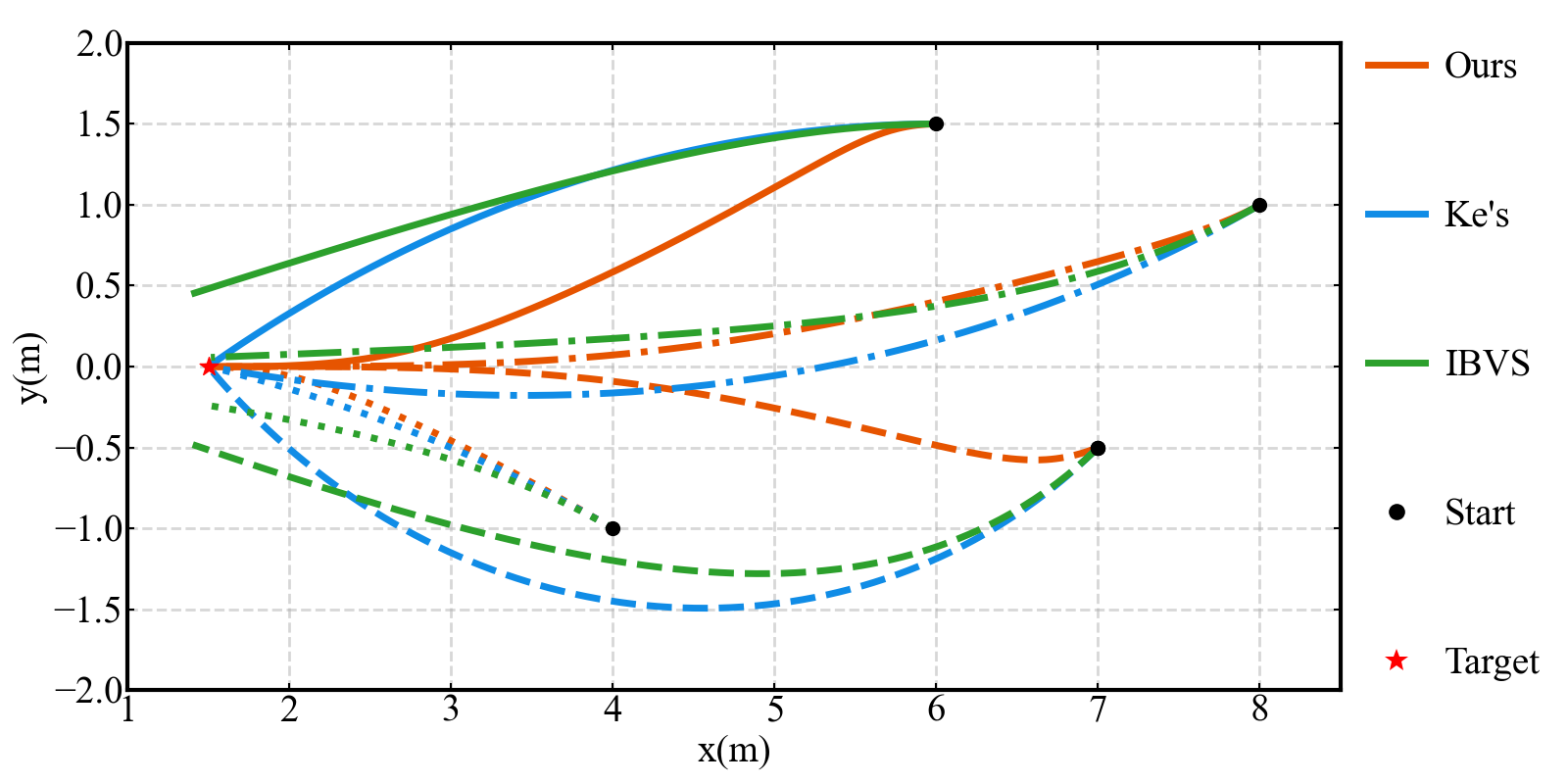}
    }\vspace{-7pt}

    \subfloat[Variation of angle with time\label{fig:curve_b}]{
        \includegraphics[width=\linewidth]{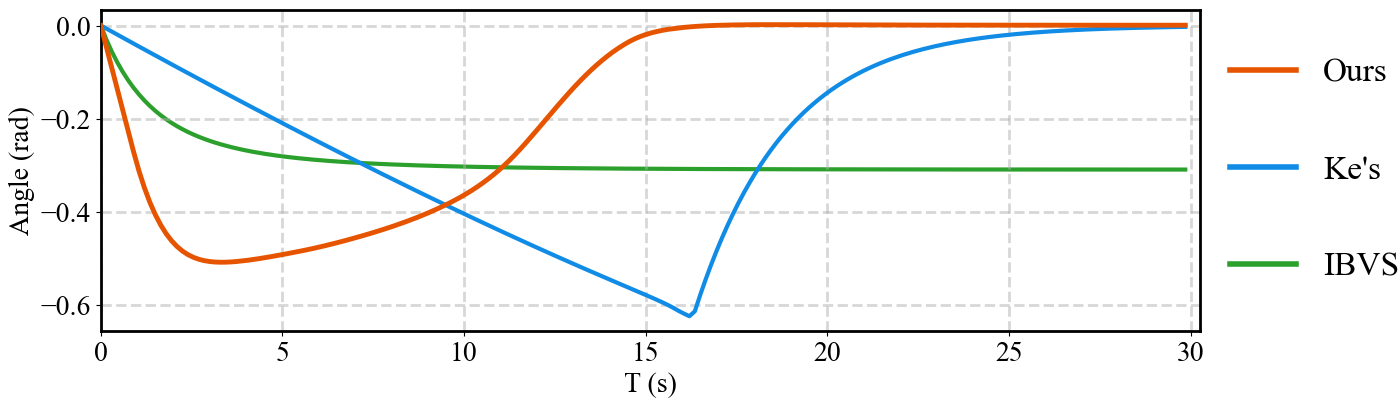}
    }\vspace{-11pt}
    
    \subfloat[Feature trajectories in image plane over time\label{fig:curve_c}]{
        \includegraphics[width=\linewidth]{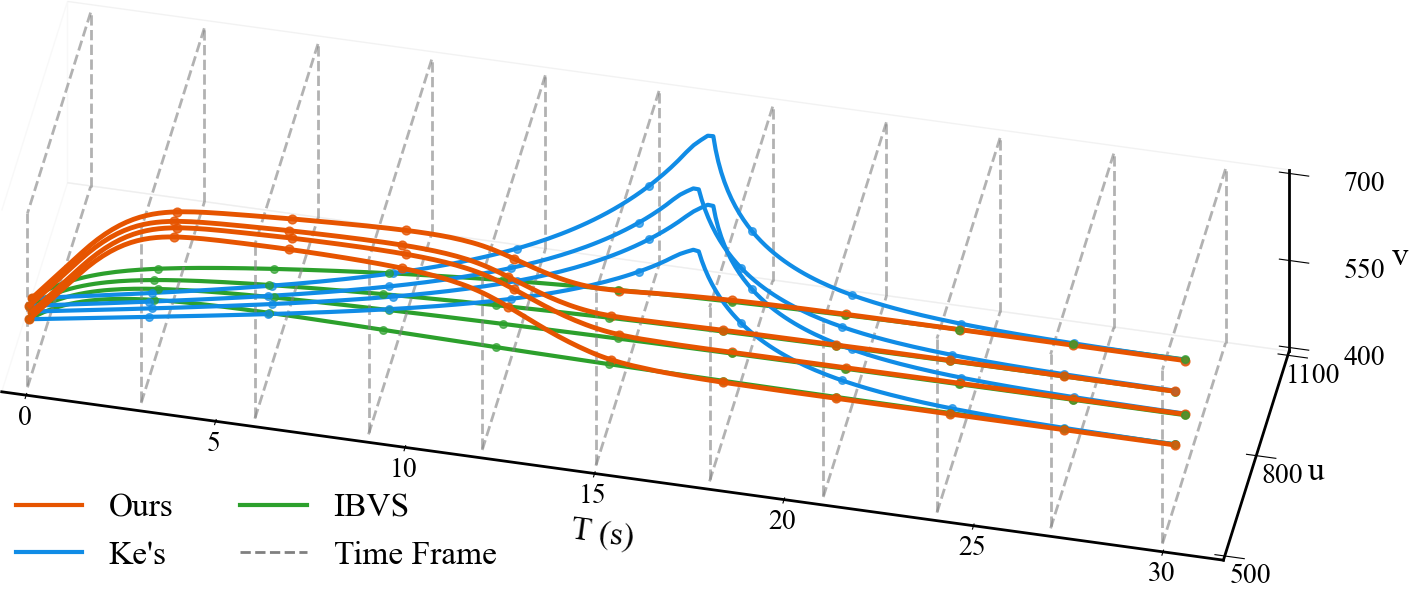}
    }\vspace{-3pt}

    \caption{Comparative simulation results.}
    \label{fig:curve}
    \vspace{-15pt}
\end{figure}

For demonstration in Fig.~\ref{fig:curve} (a), we selected four distinct initial states: $[x_F^L, y_F^L, \theta_F^L] = [6\,\text{m}, 1.5\,\text{m}, 0^\circ]$, $[8\,\text{m}, 1\,\text{m}, -30^\circ]$, $[7\,\text{m}, -0.5\,\text{m}, -45^\circ]$ and $[4\,\text{m}, -1\,\text{m}, 45^\circ]$. Three comparative trajectories were generated for each initial state, with each simulation lasting 30 seconds. In the figure, trajectories from the same initial state share a line style, while those from the same method share a color.
Focusing on the trajectories from a representative initial condition $[6\,\text{m}, 1.5\,\text{m}, 0^\circ]$, Fig.~\ref{fig:curve} (b) and (c) depict the time evolution of key state variables, including $\theta_L^F(t)$ and the feature parameters $\boldsymbol{s}(t)$.

To assess robustness under realistic sensing conditions, we injected Gaussian sensor noise with $\sigma(\boldsymbol{p}) = [3\,\text{pixel}, 3\,\text{pixel}]^\top$ for feature points and $\sigma(\theta_L^F) = \pi/60$\,rad for relative orientation. We conducted simulations from 100 randomly selected initial positions within the feature points' visible interval. Table~\ref{table_1} presents the results, comparing the metrics averaged over multiple trials between the proposed method and the baseline methods.

Experiments reveal the limitations of the baseline methods.
First, the basic IBVS approach drives the robot by aligning feature points with their desired values. However, due to the nonholonomic constraints of the platform, it cannot generate appropriately decoupled angular velocity commands. As a result, orientation errors accumulate progressively, as shown in Fig.~\ref{fig:curve} (b), eventually causing noticeable steady-state pose deviations.
Ke’s method alleviates some of these issues and is able to reach the desired final pose. Yet, as illustrated in Fig.~\ref{fig:curve} (a), it tends to prioritize reducing the position error before correcting orientation. This sequential strategy produces substantial drift in the image features (Fig.~\ref{fig:curve} (c)), which increases the likelihood of losing features from the camera FOV. In addition, the resulting curved trajectory on the XOY plane is unfavorable for a stable docking maneuver.
In contrast, the proposed method proactively regulates orientation to achieve smooth docking trajectories, demonstrating its capability to handle nonholonomic constraints. Consequently, as evidenced in Table~\ref{table_1}, it simultaneously satisfies the docking accuracy requirements and maintains smooth, well-behaved control commands.
\begin{figure}[!t]
    \centering
    \subfloat[Docking trajectories in XOY Plane]{
        \includegraphics[width=1\linewidth]{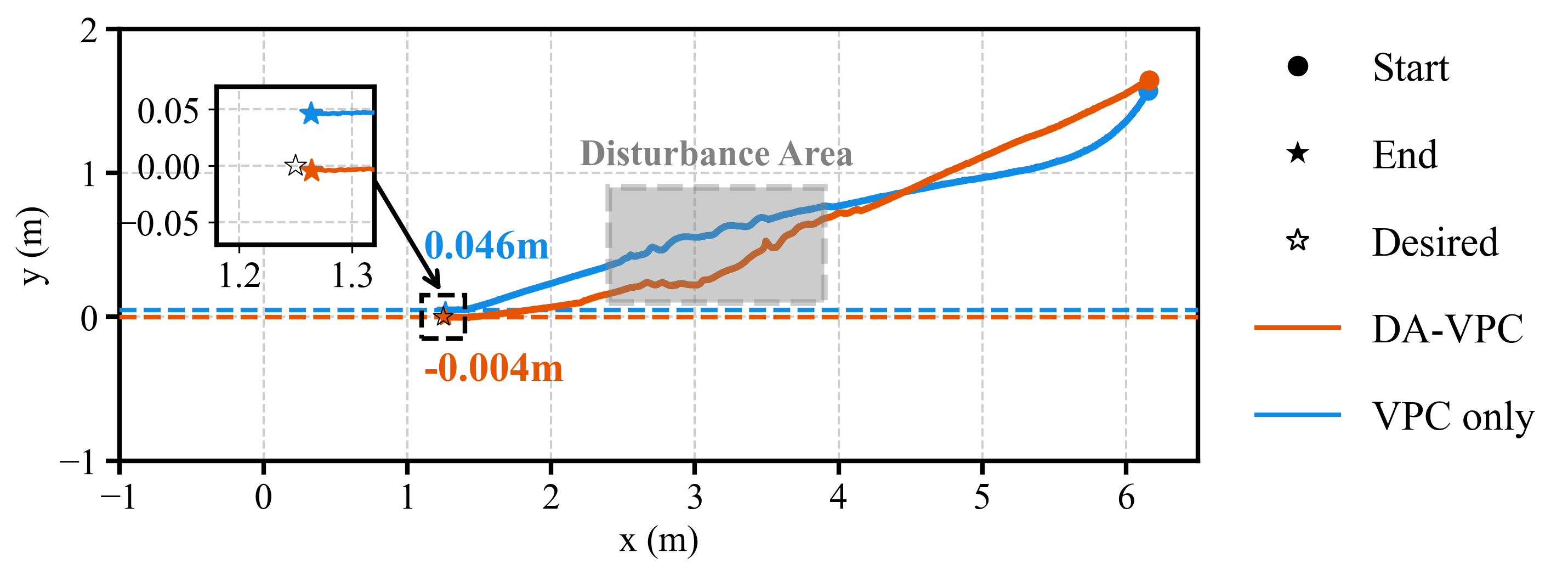}
    }
    \vspace{-5pt}
    \subfloat[Variation of disturbance estimation and compensation output over time]{
        \includegraphics[width=1\linewidth]{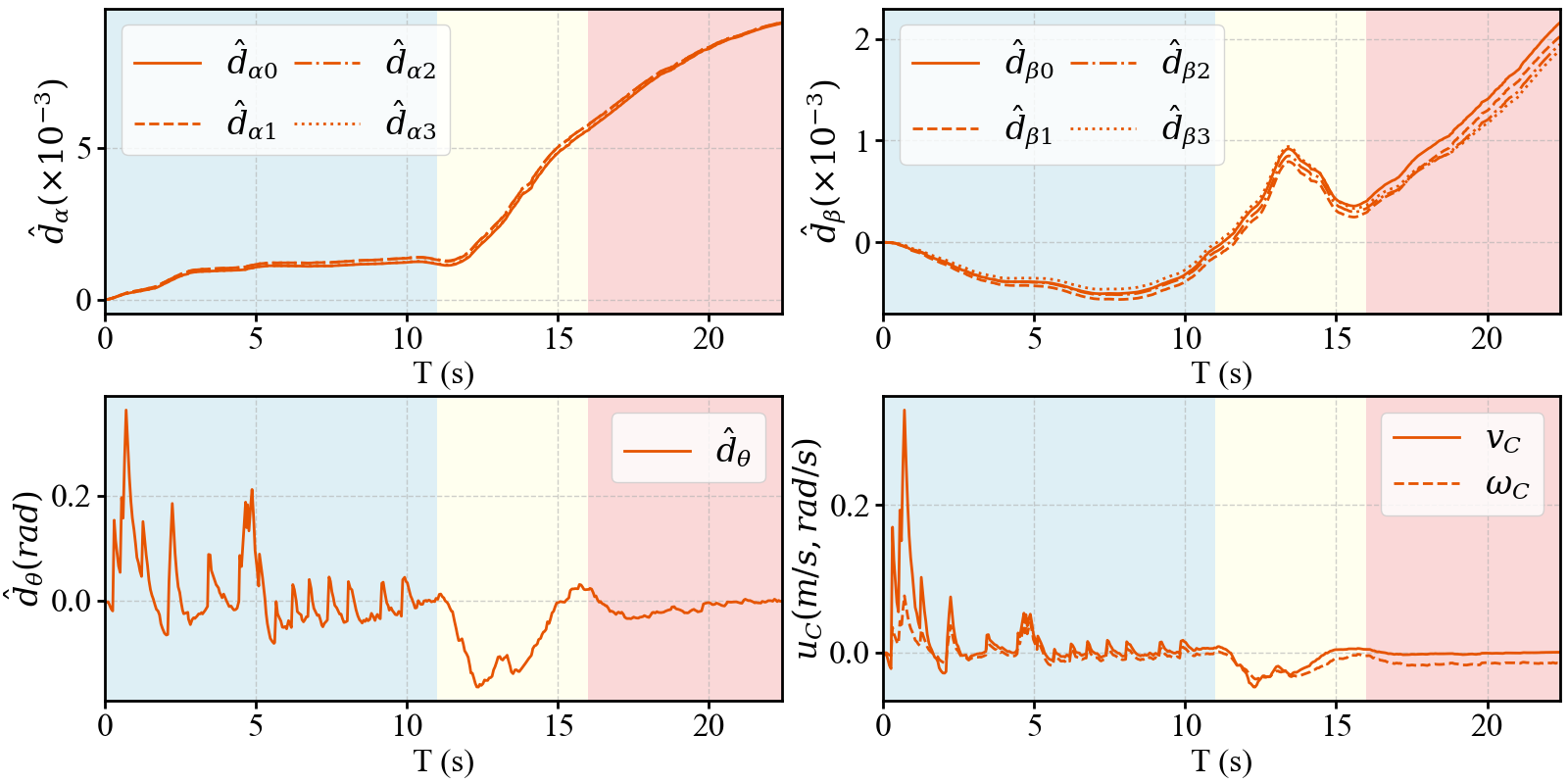}
    }
    \vspace{-5pt}
    \subfloat[Snapshots of docking maneuver under disturbed conditions]{
        \includegraphics[width=1\linewidth]{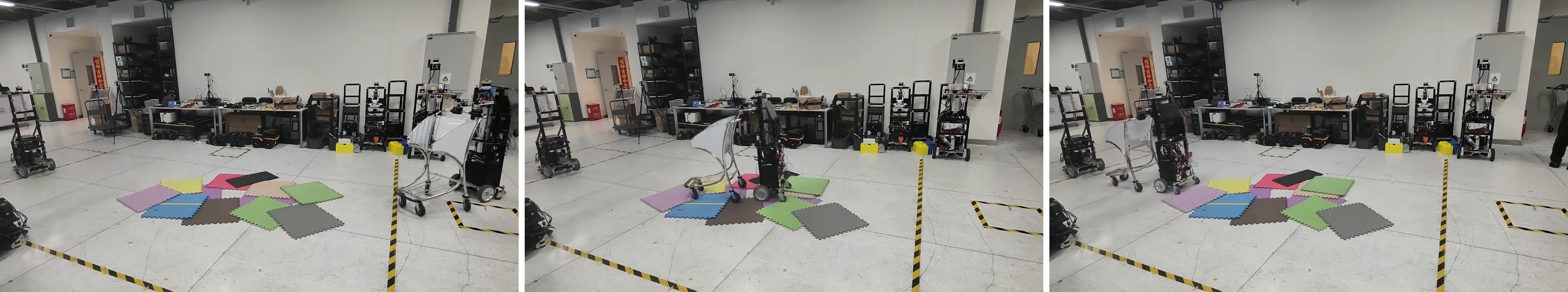}
    }
    \caption{Validation experiments of the disturbance observer. (a) shows the docking trajectory and accuracy differences with and without the disturbance rejection strategy. (b) presents the estimated values from the observer and the compensation outputs of the control inputs. (c) displays the experimental setup and three representative docking phases.}
    \label{fig:eso}
\vspace{-5pt}
\end{figure}
\begin{table}[!t]
    \renewcommand{\arraystretch}{1}
    \setlength{\tabcolsep}{2.0pt}
    \caption{Simulation Performance Comparison}
    \vspace{-0.2cm}
    \centering
    \label{table_1}
    \begin{tabular}{*{7}{c}}
    \toprule
        \multirow{3}*{Method} & 
        \multicolumn{2}{c}{$\boldsymbol{E_{p}}\downarrow$} & \multicolumn{2}{c}{$\boldsymbol{E_{s}}\downarrow$} & \multicolumn{2}{c}{$\boldsymbol{M_{sm}}\downarrow$}\\
        \cmidrule(lr){2-3} \cmidrule(lr){4-5} \cmidrule(lr){6-7}
        & $e_x$[cm] & $e_y$[cm] & $e_\theta$[degree] & $e_p$[pixel] & $v$[m/s] & $\omega$[rad/s]\\
        \midrule
        IBVS & 5.410 & 31.827 & 12.375 & 1.834 & 0.011 & \textbf{0.003} \\
        Ke's & 2.604 & 0.764 & 4.559 & 44.599 & 0.064 & 0.204\\
        Ours & \textbf{1.431} & \textbf{0.598} & \textbf{0.240} & \textbf{1.476} & \textbf{0.010} & 0.057\\
    \bottomrule
    \end{tabular} 
    \vspace{-15pt}
\end{table}
\subsection{Real-World Experiments}
\subsubsection{Disturbance Rejection Experiment}
\begin{figure*}[!t]
    \vspace{10pt}
    \centering
    \includegraphics[width=1\textwidth]{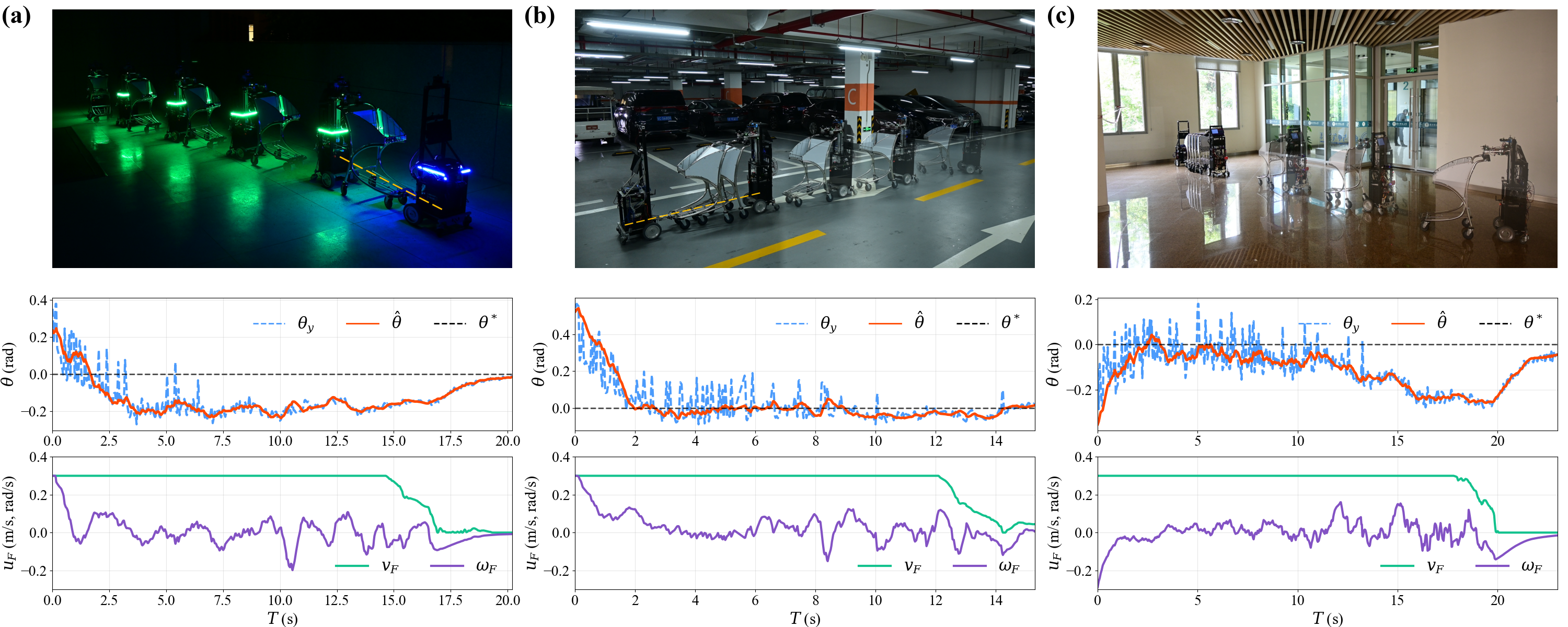}\vspace{-5pt}
    \caption{Deployment results under varying environmental conditions: (a) dark outdoors, (b) underground parking lot with lighting variation, and (c) interior featuring glass wall. Collected data includes snapshots, angle variation (measured value $\theta_y$ and estimated value $\hat{\theta}$) and control input $\boldsymbol{u_F}$.}
\label{fig:experiments}
\vspace{-15pt}
\end{figure*}
\begin{figure}[!t]
    \centering
    \includegraphics[width=0.5\textwidth]{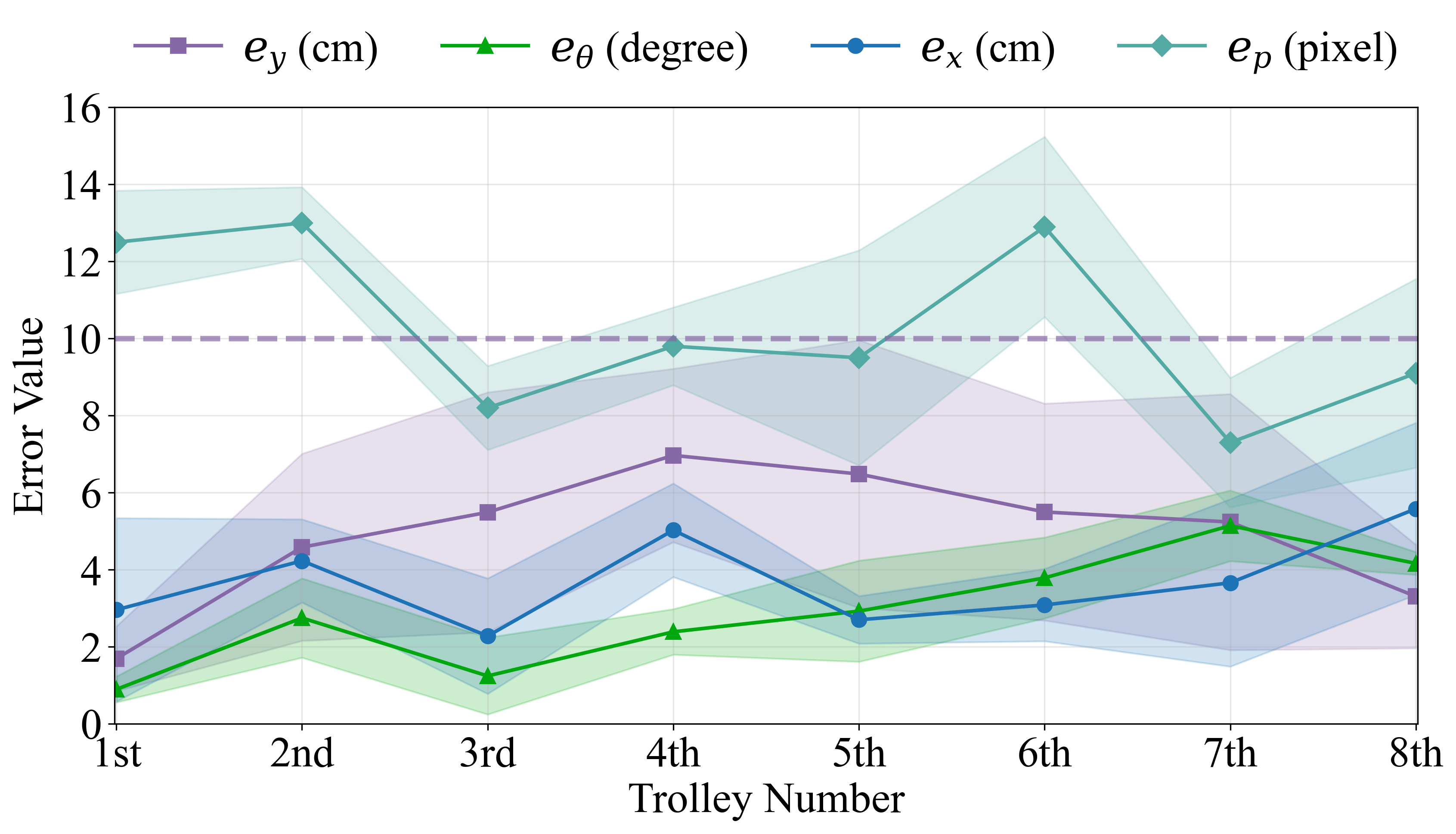}
    \caption{Line chart of the error metrics across different docking tasks.}
\label{fig:error}
\vspace{-15pt}
\end{figure}
To validate the disturbance estimation and compensation performance of the proposed method, a set of experiments was conducted to compare the docking accuracy between visual predictive control based solely on the nominal kinematic model (VPC only) and visual predictive control with disturbance estimation and compensation capabilities (DA-VPC). The initial state of the Follower robot was set to $[x_F^L, y_F^L, \theta_F^L] = [6.2\,\text{m}, 1.6\,\text{m}, -45^\circ]$. Randomly distributed rubber pads were placed in the docking trajectory area to introduce unmodeled disturbances to the system. During the experiments, which were conducted in a 6\,m × 6\,m area equipped with a motion capture system (MOCAP, 0.1\,mm accuracy) for real-time tracking, the trajectories of the Follower robot were captured and transformed into the Leader frame $\{L\}$, as shown in Fig.~\ref{fig:eso} (a). The disturbance quantities $\hat{\boldsymbol{d}}$ estimated by the disturbance observer and the compensated control input $\boldsymbol{u_C}$ are presented in Fig.~\ref{fig:eso} (b), while Fig.~\ref{fig:eso} (c) displays snapshots of the experimental process.

As illustrated in Fig.~\ref{fig:eso} (c), the docking experiment of DA-VPC is analyzed through three representative phases. These phases are distinguished in Fig.~\ref{fig:eso} (b) using blue, yellow, and red colors. In the first phase (0\,s--11\,s), due to the robot's substantial turning maneuver and angle estimation errors caused by the long distance, both $\hat{d}_{\theta}$ and $\boldsymbol{u_C}$ exhibit spike-like fluctuations. In the second phase (11\,s--16\,s), as the robot traversed the rubber pads, the observer successfully captured and estimated the significant disturbances affecting the image points and angle, with $\boldsymbol{u_C}$ actively compensating to thereby mitigate their impact. In the third phase (16\,s--22\,s), since the robot had exited the disturbance region, both $\hat{d}_{\theta}$ and $\boldsymbol{u_C}$ converged to zero. Throughout the process, it is observed that the disturbance in the image points gradually increased over time. Owing to the pinhole camera model, as the camera moves closer to the feature points, the rate of change of the image points becomes more susceptible to disturbances; nevertheless, the image disturbance remained within an acceptable range. Ultimately, the lateral docking error $e_y$ for DA-VPC reached 0.4\,cm, while the VPC only method achieved only 4.6\,cm, demonstrating the effectiveness of the proposed disturbance estimation and compensation strategy.

\subsubsection{Sequential Docking Experiment}
To evaluate the docking accuracy, the Follower robot sequentially docked with each of the eight trolleys, performing 10 trials per trolley for a total of 80 trials from randomized initial configurations.
The trajectories of the Follower robot were recorded by MOCAP to assess docking errors. Fig.~\ref{fig:error} presents the line charts of docking errors across all 8 experimental sets.

For the first trolley, the critical docking accuracy metrics—lateral position error $e_y^{(1)}$ and angular alignment error $e_\theta^{(1)}$—achieved mean values of $1.69\,\text{cm}$ and $0.84^\circ$, respectively, both well within the prescribed bounds
$
e_y^{(1)} \le \bar{e}_y^{(1)} = 6~\text{cm}.
$
As additional trolleys were docked $(k>1)$, the expected feature images became progressively smaller—thereby increasing the docking difficulty—and both $e_y^{(k)}$ and $e_\theta^{(k)}$ exhibited a gradual increasing trend. Nevertheless, all lateral errors remained below the required upper bound.
$
e_y^{(k)} \le \bar{e}_y^{(k)} = 10~\text{cm},
$
as indicated by the dashed line, confirming that the proposed method consistently satisfies the docking accuracy requirements for all trolleys.

\subsubsection{Multi-scenario Experiment}

A comprehensive validation of the integrated hardware-software system performance is conducted through three sets of trolley docking experiments in diverse real-world scenarios. The snapshots and corresponding data are presented in Fig.~\ref{fig:experiments}. Note that no sensors other than the proposed vision system are employed in experiments.

In dark outdoor environments (Fig.~\ref{fig:experiments} (a)), non-active visual features like AprilTag cannot be detected and extracted. 
An initial lateral error of approximately 2\,m was intentionally set for docking the first trolley.
After 2\,s, the orientation error gradually increases and then converges to zero, which matches the simulation results and demonstrates the capability of our method in handling nonholonomic constraints.
The second docking experiment is conducted in an underground parking lot (Fig.~\ref{fig:experiments} (b)). The system maintained successful docking performance for the second trolley despite dynamically flickering lighting and irregular terrain surfaces.
Finally, the docking experiment of the fifth trolley is performed near the sunlight-transmitting windows, reflective floor and glass partitions (Fig.~\ref{fig:experiments} (c)), starting from an initial distance of 8\,m. It demonstrates that the system maintains stable operation despite environmental light interference.
The experimental data indicate that when two robots are at a long distance, the detection module exhibits significant noise in angle measurement. However, the system demonstrates strong resilience to angular fluctuations, confirming the effectiveness of the proposed disturbance rejection strategy.

\section{Conclusions}

This paper presents a monocular vision-based robotic docking system for reliable trolley collection in complex environments.
By integrating active infrared features with visual predictive control, our approach achieves millimeter-level docking accuracy while addressing the challenges of nonholonomic constraints, environmental variability and disturbances. 
We validate the system in diverse real-world settings, including dark outdoor areas and underground parking lots under various lighting and layout conditions. 
This work advances the practical deployment of autonomous logistics robots in unstructured public spaces.
Although the IR-based method ensures accurate extraction of target features, it inherently limits perception of the external environment, necessitating further extensions for dynamic obstacle avoidance.
Future work will explore vision-based whole-body control and cooperative trolley transportation system, leveraging accelerated computing for improving real-time performance.

\bibliographystyle{IEEEtran}
\bibliography{reference}

\end{document}